\documentclass[letterpaper,10pt,conference]{ieeeconf}
\IEEEoverridecommandlockouts                              
\usepackage{graphicx}      
\usepackage[ruled,vlined]{algorithm2e}
\SetNlSty{bfseries}{\color{black}}{}
\usepackage{amsmath}
\usepackage{amssymb}
\usepackage{soul,color}
\newcommand{\norm}[1]{\left\lVert#1\right\rVert}

\usepackage{pgfplots}
\usepackage{tikz}
\usetikzlibrary{external,calc,patterns,decorations.pathmorphing,decorations.markings,decorations.text,arrows,shapes,positioning, backgrounds}
\definecolor{ffqqtt}{rgb}{1,0,0.2}
\definecolor{qqqqff}{rgb}{0,0,1}
\pgfplotsset{width=10cm,compat=1.9}
\usepackage[utf8x]{inputenc} 

\usepackage{caption}
\usepackage{subcaption}

\newcommand\copyrighttext{%
  \footnotesize \copyright 2021 IEEE.  Personal use of this material is permitted.  Permission from IEEE must be obtained for all other uses, in any current or future media, including reprinting/republishing this material for advertising or promotional purposes, creating new collective works, for resale or redistribution to servers or lists, or reuse of any copyrighted component of this work in other works.}
\newcommand\copyrightnotice{%
\begin{tikzpicture}[remember picture,overlay]
\node[anchor=south,yshift=10pt] at (current page.south) {\fbox{\parbox{\dimexpr\textwidth-\fboxsep-\fboxrule\relax}{\copyrighttext}}};
\end{tikzpicture}%
}

\title{\LARGE \bf COLREGs-Informed RRT* for Collision Avoidance of Marine Crafts}

\author{Thomas Thuesen Enevoldsen$^{1}$, Christopher Reinartz$^{1}$ and Roberto Galeazzi$^{1}$
\thanks{*This research is sponsored by the Danish Innovation Fund, The Danish Maritime Fund, Orients Fund and the Lauritzen Foundation through the Autonomy part of the ShippingLab project.}
\thanks{$^{1}$Automation and Control Group, Department of Electrical Engineering, Technical University of Denmark, DK-2800 Kgs. Lyngby, Denmark
        {\tt\small \{tthen,ccrein,rg\}@elektro.dtu.dk}}%
}

\begin{document}

\maketitle

\thispagestyle{empty}
\pagestyle{empty}

\copyrightnotice
\begin{abstract}
The paper proposes novel sampling strategies to compute the optimal path alteration of a surface vessel sailing in close quarters. Such strategy directly encodes the rules for safe navigation at sea, by exploiting the concept of minimal ship domain to determine the compliant region where the path deviation is to be generated. The sampling strategy is integrated within the optimal rapidly-exploring random tree algorithm, which minimizes the length of the path deviation. Further, the feasibility of the path with respect to the steering characteristics of own ship is verified by ensuring that the position of the new waypoints respects the minimum turning radius of the vessel. The proposed sampling strategy brings a significant performance improvement both in terms of optimal cost, computational speed and convergence rate.
\end{abstract}

\section{Introduction}
The transport sector is witnessing an increasing interest in the adoption of autonomous systems' technologies, driven by the expectation of e.g. improving the intermodal logistic chain and solving the last mile problem \cite{ranieri2018review,oliveira2017sustainable}. The deployment of autonomous means of transport such as busses, trucks and ships strongly depends on the development of two key technologies: Situation awareness and collision avoidance. Despite the technological needs are common across the different transport domains, the technologies are not easily portable due to the specificity of the operational environment and regulations that dictate how vehicles must operate. 
The pathway towards full autonomous operations is convoluted since social, economic and normative barriers need to be addressed alongside the technical challenges \cite{komianos2018autonomous,kim2020autonomous}. However, the intermediate adoption of technologies providing partly autonomous capabilities integrated within decision support systems seems in clear sight, as already shown by the advanced driving assistance systems in the automotive industry. 
A similar focus is present in the maritime cluster that expects advanced navigation assistance systems (ANAS) to be one of the first byproducts of the autonomous ship development. ANAS will encompass both situation awareness and collision avoidance systems; however the decision making process will remain rooted within the ship navigator. The introduction of ANAS in merchant vessels is expected to increase the operational safety, thereby reducing the likelihood of incidents caused by human error, which is still one of the main factors~\cite{emsa2019}.

This paper addresses the design of the short horizon planner for collision avoidance. The essence of collision avoidance for marine vessels revolves around finding a feasible path deviation from the nominal planned route to avoid potential collision and grounding situations, whilst adhering to the International Regulations for Preventing Collisions at Sea (COLREGs). The paper proposes novel sampling strategies for the optimal rapidly-exploring random tree (RRT*) algorithm, which directly encodes the COLREGs by defining a compliant region of the configuration space where feasible paths should be searched for.

\subsection{Literature review}
RRT-based algorithms have been widely adopted by the robotics community since it was first proposed in \cite{lavalle1998rapidly,LaValle2001}. To address the lack of optimality of the found path, \cite{karaman2010incremental,karaman2011sampling} proposed RRT*, which allows the inclusion of optimization metrics to improve the quality of the obtained solutions as the number of samples goes to infinity, thus ensuring asymptotic optimality. To further improve the performance of RRT* \cite{Gammell2014} proposed Informed-RRT*, which constrains the search for the optimal solution in an elliptical region of the configuration space once a feasible path is found.

RRT-based algorithms have also been applied for the design of collision avoidance schemes for marine vessels. For explanatory details regarding specific COLREGs, see Section~\ref{sec:COLREGs}. In \cite{Chiang2018} a non-holonomic RRT (RRT* \cite{Park2015,Pharpatara2017}) is combined with an aggressive goal-seeking strategy to generate a path that complies with COLREGs rules 14-15. Compliance is achieved by describing the target vessels as virtual obstacles, which extends their collision boundaries, and forces the tree to grow around a given target vessel. \cite{zaccone2018random,Zaccone2020} implemented an RRT* algorithm that produces a tree of waypoints optimized with respect to distance, curvature and repulsion from obstacles. Later in~\cite{Zaccone2019} an extension is made to include the COLREGs, where the RRT* algorithm will reject non COLREGs compliant samples.

Research on collision avoidance for autonomous ships has been extensive in recent years, and~\cite{huang2020ship} provides a broad survey of the developed algorithms.
\cite{HU201713662,HU2019Naeem} proposed a multi-objective approach for the generation of COLREGs compliant paths using the human element of \emph{good seamanship}, which was assessed based on expert knowledge. \cite{NAEEM2012669} proposed a collision avoidance scheme that is compliant with COLREGs rules 8 and 14, with the scheme generating waypoints adhering to the vessel dynamics whilst considering both static and dynamic objects. \cite{NAEEM2016207} detailed a modified version of the Artificial Potential Fields algorithm, which generates COLREGs compliant paths (adhering to rules 13, 14 and 15), whilst considering a multi-vessel encounter with static obstacles. Model Predictive Control (MPC) has also been adopted for the design of collision avoidance schemes. \cite{eriksen2017mpc} showed an implementation addressing both static and dynamic obstacles, which minimizes the deviation from the nominal course. Later, \cite{ERIKSEN2019} presented the branching-course MPC that favours trajectories that are compliant with rules 14 and 15, whilst adhering to rules 8, 13 and 17. 
\cite{benjamin2004colregs} and \cite{benjamin2006navigation} presented a detailed account of how to represent the COLREGs in an autonomous setting, and proposed a multi-objective optimization method within a behaviour-based control framework to obtain compliant traversal.
Various variants of popular search-based algorithms have also been investigated in the marine setting: \cite{SINGH2018187} used the A* algorithm to compute a collision free path in the presence of ocean currents; \cite{CAMPBELL2012386} investigated a heuristic rule-based A* (R-RA*) algorithm to develop a real time COLREGs compliant path planner, which was validated on a bridge simulator \cite{CAM2014naeem}. Other collision avoidance schemes include approaches using genetic algorithms \cite{ni2018modelling,ni2019}, Ant Colony Optimization \cite{Lazarowska_2015,Lazarowska2019} and lattice-based methods \cite{Shah2016}.

\subsection{Novelty and contribution}
The paper proposes two new sampling strategies for the RRT* algorithm to naturally translate the COLREGs into geometrical regions of the configuration space, where feasible and collision free paths have to be generated. Exploiting the concept of minimal ship domain, half-annulus and elliptical half-annulus regions are proposed as search areas for the path alteration, and corresponding sampling schemes are derived to ensure uniform coverage of such regions. This implicitly defines a-priory forbidden zones that RRT* cannot explore because of breaching the COLREGs. The paper shows that the proposed sampling strategies significantly improve the performance of the RRT* algorithm with respect to the metrics of mean time to discover a feasible path, computational speed to find the optimal solution and convergence to the optimal cost, when compared to implementations adopting reference sampling strategies and rejection sampling \cite{karaman2011sampling,Zaccone2020,Gammell2014}.

\section{Preliminaries}
\subsection{Vessel and control scheme assumptions}
This study focuses on path planning for merchant vessels, such as ferries, ro-ro vessels and container feeders, whose maneuvering capability is described by the minimum turning radius $R_{min}$.

The considered collision scenarios are assumed to occur during open water passage with single vessel encounters. In such situations, human navigators avoid collision by minimizing the path alteration and trying to avoid speed variations. Therefore, it is further assumed that both own ship and target vessel will proceed with constant speed throughout the collision scenario.
Moreover, the steering dynamics of own ship is controlled through an autopilot or track pilot, which receives a sequence of waypoints $W_i = (N_i,E_i,R_i)$, where $N_i$ is the North position, $E_i$ is the East position, and $R_i$ is the radius of acceptance, which are feasible with respect to the manoeuvring capabilities (i.e minimal turning radius). Given the radius of acceptance $R_i$ it is possible to compute the turning radius $\bar{R}_i$ required to transition between two consecutive legs of the path, i.e. 
\begin{equation}
    \bar{R}_i = R_i\tan\left(\frac{\phi}{2}\right), \quad \phi = (\alpha - \beta  + \pi) \mod 2\pi\label{turningrad_req}
\end{equation}
where $\alpha$ and $\beta$ describe the path tangential angles of the two legs. Consecutive waypoints must be positioned such that $\bar{R}_i \geq R_{min}$.

\subsection{Target vessel representation}
Using AIS (Automatic Identification System) data from vessels sailing in Southern Danish waters \cite{hansen2013empirical} estimated the minimum ship domain in which a navigator feels comfortable. Such comfort zone can be described by an ellipse with major axis $a_{sd} = 8L$ and minor axis $b_{sd} = 3.2L$, where $L$ is the ship length. The minimum ship domain can be utilized within the short horizon planner to ensure the generation of a collision free path between own ship and the target vessels. 

At time $t = \bar{t}$, a target vessel violates the comfort zone of own ship if the following inequality is true
\begin{equation}
\begin{split}
&\frac{\left(\Delta E(\bar{t})\sin \psi(\bar{t}) + \Delta N(\bar{t})\cos \psi(\bar{t})\right)^{2}}{\left(\frac{a_{sd}}{2}\right)^{2}}+\\
&\frac{\left(\Delta E(\bar{t})\cos \psi(\bar{t}) - \Delta N(\bar{t})\sin \psi(\bar{t})\right)^{2}}{\left(\frac{b_{sd}}{2}\right)^{2}} \leq 1
\end{split}
\end{equation}
where $\Delta E(\bar{t}) = E_{OS}(\bar{t})-E_{TV}(\bar{t}) $ and $\Delta N(\bar{t}) = N_{OS}(\bar{t})-N_{TV}(\bar{t})$ are the difference in East and North directions of own ship and the target vessel, and $\psi(\bar{t})$ is the heading of the target vessel.

Last, it is assumed that position, speed and heading of the target vessel are available through e.g. AIS or RADAR.

\subsection{COLREGs overview}\label{sec:COLREGs}

\begin{figure}[tb]
\centering
\includegraphics[width=0.4\textwidth]{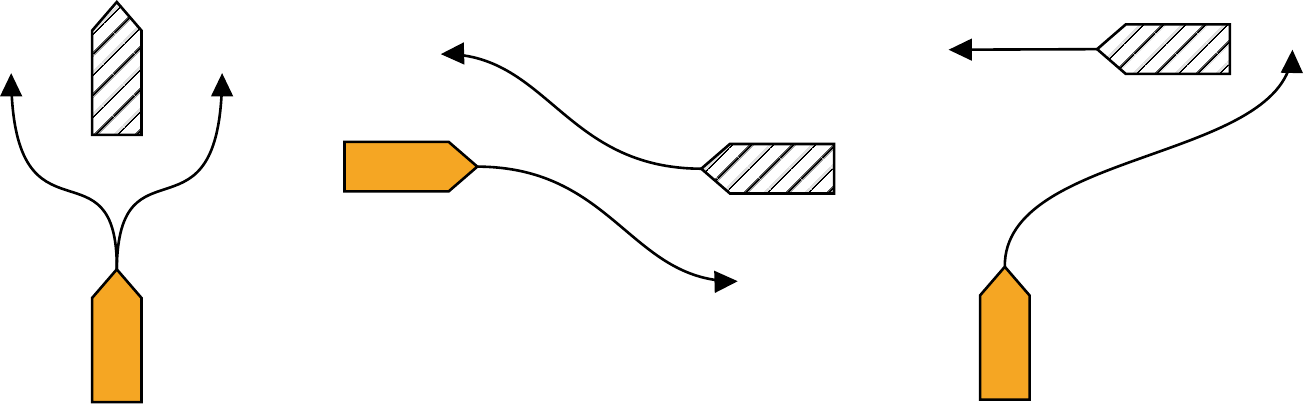}
\caption{Visualization of the encounters described by COLREG rules 13-15, i.e. overtaking, head-on and crossing scenarios. The solid coloured vessel is in a give-way situation.}\label{R13_15}
\vspace{-0.5cm}
\end{figure}
The following is a brief overview of the International Regulations for Preventing Collisions at Sea (COLREGs, \cite{imo1972}), which an advanced navigation assistance system must abide by. The listed rules describe the most commonly occurring collision risk scenarios for two vessels, and only apply if both vessels are power-driven. A graphical representation of rules 13-15 is provided in Fig.~\ref{R13_15}.\\
\textit{Rule 13, Overtaking:} The vessel being overtaken must maintain current course and speed, whereas the other may overtake on either side.\\
\textit{Rule 14, Head-on:} Both vessels must perform a manoeuvre such that they pass one-another on their port sides.\\
\textit{Rule 15, Crossing:} A vessel must give-way to another vessel, if the second vessel approaches from starboard side. The vessel that has the right of way must keep the current course and speed.\\
\textit{Rules 16-17:} Specify "good behaviour" of a given vessel, depending on whether it is the stand-on or give-way vessel. Rule 16 dictates that the give-way vessel should take early action and remain a safe distance from the stand-on vessel. Rule 17 urges a vessel in the stand-on situation to remain so, and not attempt to avoid collision unless it is clear that the give-way vessel is not abiding by the COLREGs. 

\subsection{CPA and TCPA}

\begin{figure}
     \centering
     \begin{subfigure}[b]{0.23\textwidth}
\centering
\resizebox{1\linewidth}{!}{
\begin{tikzpicture}[line cap=round,line join=round,>=triangle 45,x=1.0cm,y=1.0cm]
\clip(-0.1,-0.1) rectangle (5,5.64);
\draw [->,dash pattern=on 2pt off 2pt] (0,0) -- (2,3);
\draw [->,dash pattern=on 2pt off 2pt] (0,0) -- (4,3);
\draw [->,color=qqqqff] (4.5,0.5) -- (4.01,0.99);
\draw [->,color=ffqqtt] (4.5,4.5) -- (4.29,3.86);
\draw [dash pattern=on 2pt off 2pt,color=ffqqtt] (4,3)-- (4.5,4.5);
\draw [dash pattern=on 2pt off 2pt,color=qqqqff] (4.5,0.5)-- (2,3);
\draw [->,color=qqqqff] (2,3) -- (1.51,3.49);
\draw [->,color=ffqqtt] (4,3) -- (3.77,2.32);
\draw [->] (0,0) -- (0,5.01);
\draw [->] (0,0) -- (5,0);
\draw (0.17,4.98) node[anchor=north west] {N};
\draw (4.65,0.6) node[anchor=north west] {E};
\draw (2,3)-- (4,3);
\fill [color=qqqqff] (2,3) circle (1.5pt);
\fill [color=ffqqtt] (4,3) circle (1.5pt);
\fill [color=ffqqtt] (4.5,4.5) circle (1.5pt);
\fill [color=qqqqff] (4.5,0.5) circle (1.5pt);
\draw[color=qqqqff] (3.8,0.61) node {$t=0$};
\draw[color=ffqqtt] (4,4.5) node {$t=0$};
\draw[color=black] (0.8,1.61) node {$\mathbf{p}_{OS}\,\,\,\,$};
\draw[color=black] (2.2,1.3) node {$\,\,\,\mathbf{p}_{TV}$};
\draw[color=qqqqff] (4.65,0.8) node {$\mathbf{v}_{OS}$};
\draw[color=ffqqtt] (4.7,4.04) node {$\,\mathbf{v}_{TV}$};
\draw[color=qqqqff] (2.08,3.25) node {$\mathbf{v}_{OS}$};
\draw[color=ffqqtt] (4.20,2.53) node {$\,\,\,\,\mathbf{v}_{TV}$};
\draw[color=black] (2.93,3.2) node {$\mathrm{CPA}$};
\end{tikzpicture}}
\caption{}\label{TCPA_CPA_fig}

     \end{subfigure}
     \hfill
     \begin{subfigure}[b]{0.23\textwidth}
         \centering
	\centering
	\resizebox{1\linewidth}{!}{\begin{tikzpicture}[shorten >=1pt,on grid,
		every initial by arrow/.style={*->}]
		\draw (2,2) -- (3,2) -- (3,4) to [out=100,in=-30,looseness=0.75241] (2.5,5) to [out=210,in=80,looseness=0.75241] (2,4) -- (2,2);
		\draw[dashed] (2.5,3) -- node[right,pos=1.0]{$\beta = 0^o$} (2.5,6.3);
		\node at (1.0,3.7) (E1){Port};
		\node at (4.125,3.7) (E2){Starboard};
		\node at (2.5,0.25) (E45){Overtaking};
		\coordinate (origin) at (2.5,3){};
		\coordinate (e1) at (2.5,6);
		\coordinate (e2) at (2.3169,5.9944);
		\coordinate (e4) at (5.2716,1.8519);
		\coordinate (e5) at (-0.2716,1.8519);
		\draw (origin) -- node[left,pos=0.75] {$\beta = -3.5^o$}(e2);
		\filldraw [draw=black, pattern=north west lines, pattern color=green,line width=0.5, dashed]
		(origin) -- (e1) arc [radius=5.2, start angle=90, end angle=92] (e2) -- (origin) -- cycle;
		\draw (origin) -- node[below,pos=0.75,yshift=-0.45cm]{$\beta = 112.5^o$} (e4);
		\draw (origin) -- node[below,pos=0.75,yshift=-0.45cm]{$\beta = -112.5^o$} (e5);
		\end{tikzpicture}}
	\caption{} \label{fig:relbearing}

     \end{subfigure}
        \caption{(a) Vector definitions for the derivation of CPA and TCPA for two vessels with constant speed and heading. (b) Angles used to determine the relative position between own ship and the given target vessel. The gridded area indicates uncertainty associated with the head-on situation, see \cite{papageorgiou2019parallel}. }
        \vspace{-0.5cm}
\end{figure}
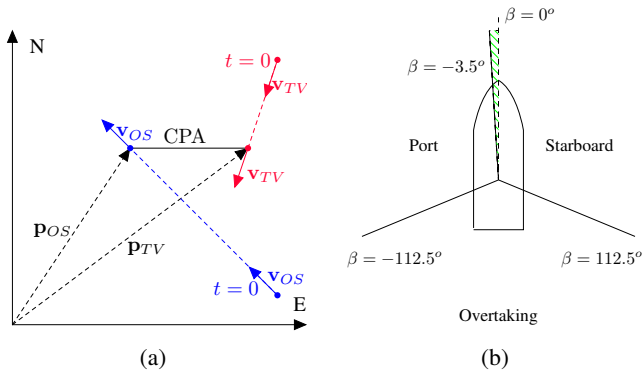

The Closest Point of Approach (CPA) and Time to Closest Point of Approach (TCPA) are two metrics used by master mariners in order to navigate a vessel in a safe manner, where CPA together with the relative bearing is a determining factor for whether or not a COLREG situation is developing, and an evasive manoeuvre must be performed.

Let $\mathbf{p}_{OS} = [N_{OS},E_{OS}]^T$ and $\mathbf{p}_{TV} = [N_{TV},E_{TV}]^T$ be the position of own ship and target vessel in the North-East inertial frame. Let $\mathbf{v}_{OS} = [V_{OS}\sin\psi_{OS},V_{OS}\cos\psi_{OS}]^T$ and $\mathbf{v}_{TV} = [V_{TV}\sin\psi_{TV},V_{TV}\cos\psi_{TV}]^T$ be the own ship and target vessel velocity vectors in the same frame, where $V_{(\cdot)}$ is the vessel's speed and $\psi_{(\cdot)}$ is the vessel's heading.  
Knowing the positions and velocities of both vessels at the current time, TCPA is defined as
\begin{equation}
\mathrm{TCPA} \triangleq - \frac{\left(
	\Delta \mathbf{p}(0)\right)^T\Delta \mathbf{v}}
{\left(\Delta \mathbf{v}\right)^T\Delta \mathbf{v}}
\end{equation}
with $\Delta \mathbf{v} = \mathbf{v}_{OS} - \mathbf{v}_{TV}$, $\Delta \mathbf{p}(0) = \mathbf{p}_{TV}(0) - \mathbf{p}_{OS}(0)$. Assuming constant speed for both vessels, their respective positions at time TCPA is computed as
\begin{align}
\mathbf{p}_{OS}(\mathrm{TCPA}) &= \mathbf{p}_{OS}(0) + \mathbf{v}_{OS} \cdot \mathrm{TCPA} \\
\mathbf{p}_{TV}(\mathrm{TCPA}) &= \mathbf{p}_{TV}(0) + \mathbf{v}_{TV} \cdot \mathrm{TCPA}
\end{align}
The CPA is then defined as follows
\begin{equation}
\mathrm{CPA} \triangleq \norm{\Delta \mathbf{p}(\mathrm{TCPA}) - \frac{\left(
		\Delta \mathbf{p}(0)\right)^T\Delta \mathbf{v}}
	{\left(\Delta \mathbf{v}\right)^T\Delta \mathbf{v}}\Delta \mathbf{v}}
\end{equation}
which provides a metric of how close the two vessels will be at $t=\mathrm{TCPA}$. A graphical interpretation of this derivation is provided in Fig.~\ref{TCPA_CPA_fig}.

\section{Sampling the COLREGs compliant subset}\label{section_creg_sampling}
\subsection{Defining the COLREGs compliant subset}\label{section_creg_samp_def}
\begin{figure}[tb]
\centering
\includegraphics[width=0.45\textwidth]{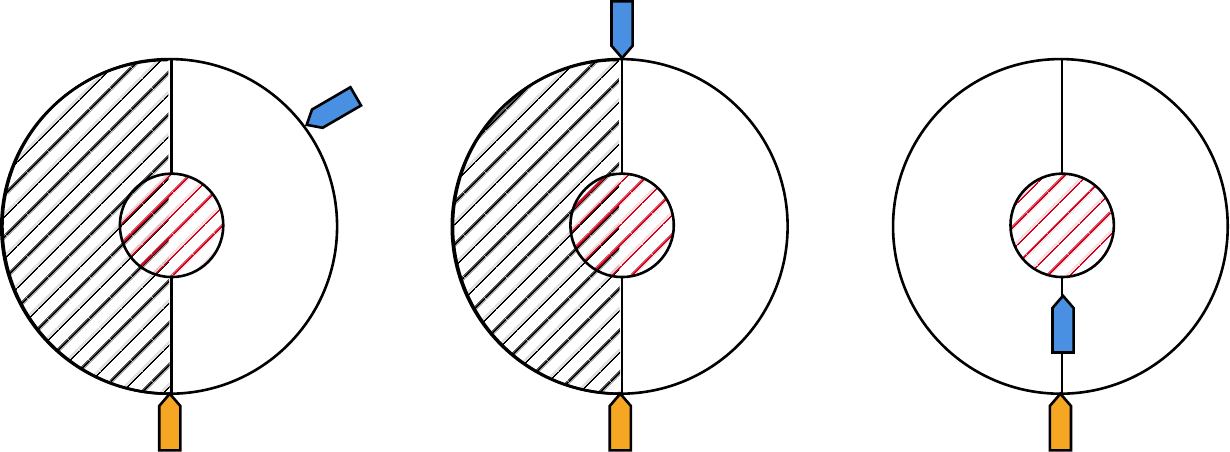}
\caption{Three COLREGs compliant subsets, given the different encounter types. The outer-circle radius is defined as the the limit $t_{act}$, with the radius of the inner circle being $d_{act}$. }\label{COLREGs_subset}
\vspace{-0.5cm}
\end{figure}
To improve the sampling performance of RRT* the COLREGs compliant sampling space is built. First, the current COLREGs scenario must be identified; this can be achieved by using a limit on the minimum allowed CPA along with the relative bearing between the vessels. If $\mathrm{CPA} < d_{act}$, where $d_{act}$ is a specified distance, then at $t=\mathrm{TCPA}$ the two considered vessels will enter too close quarters and either collide or pose a collision risk. The relative bearing is the determining factor for which vessel is in the stand-on or give-way situation. Fig.~\ref{fig:relbearing} details the various relative angles, which provides information regarding own ships location compared the target vessel. If OS is located on the port-side of the TV, then OS is the give-way vessel, and must therefore perform a COLREGs compliant manoeuvre to mitigate the risk of collision. This manoeuvre is initiated when $\mathrm{TCPA} < t_{act}$ where $t_{act}$ is a specified time. Both $d_{act}$ and $t_{act}$ strongly depend on the vessel type, and typically increase in size as the vessel size increases.

The single vessel encounters for rules 13-15 (Fig.~\ref{R13_15}), can be represented in terms of two circles, as shown in Fig.~\ref{COLREGs_subset}, with the radius of the inner circle determined by $d_{act}$ and the outer radius by $t_{act}$. Navigating within the inner circle violates the CPA limit, and the outer circle represents the area (given the time span) where action is required. The plain areas represent points in time which are COLREGs compliant with respect to the identified situation, assuming that the stand-on vessel maintains constant course and speed. The sampling space can be represented as a half-annulus for all three scenarios, with the overtaking scenario having the freedom to be represented as an annulus. By limiting sampling to this space, it is ensured that all points are valid with the respect to the COLREGs, rendering COLREGs specific rejection sampling unnecessary. 

The notion of this COLREGs compliant subspace can be further extended by including ideas from the sampling techniques presented in \cite{Gammell2014}, namely the Informed-RRT*. The authors argued that the probability of improving the overall cost of the found solution is low, whilst continuing to sample in the original sampling space since it contains an abundance of points that do not improve the solution. The strategy of using the COLREGs compliant subset is extended such that once an initial solution is obtained using the half-annulus sampling space, a second space is introduced. This second space is the elliptical half-annulus, which has dimensions similar to the informed subset introduced in~\cite{Gammell2014}. A notable difference is that the informed subset uniformly samples an entire ellipse, whereas the COLREGs compliant subset consists of an elliptical half-annulus. The length of the ellipse is equal to the current cost of the best solution $c_{best}$ and the width is equal to $\sqrt{c_{best}^2 - c_{min}^2}$, where $c_{min}$ is the straight line distance between the start and end nodes.

\subsection{Uniform sampling of the half-annulus subset}\label{samplinghalfannulus}

In order to uniformly sample the annulus, we apply the inversion method for non-uniform random variate generation~\cite[Chapter 2]{devroye1986non}. Let $X$ be a random variable with probability density function (PDF)
\begin{equation}
f(x) \triangleq \underbrace{\frac{A_{\text{inner}}}{A_{\text{circle}}}}_{c_i} +2 \underbrace{\frac{A_{\text{outer}}}{A_{\text{circle}}}}_{c_o}x
\end{equation}
where $A_{\text{inner}} = \pi r_{min}^2$, $A_{\text{outer}} = \pi \left(r_{max}^2 - r_{min}^2\right)$ and $A_{\text{circle}} = \pi r_{\text{max}}^2$, with inner radius $r_{min}$ and an outer radius $r_{max}$. The cumulative distribution function (CDF) is then defined as,
\begin{equation}
F(x) \triangleq c_i x + c_o x^2
\end{equation}
and its inverse can be found by solving the equality $F\left(F^{-1}\left(u\right)\right) = u$ for $F^{-1}(u)$ where $u \sim \mathcal{U}(0,1)$ is a realization of the uniform distribution. In the original set of parameters $F^{-1}(u)$ reads
\begin{equation}
\begin{split}
F^{-1}(u) &= \frac{-r_{min}^2}{2\left(r_{max}^2-r_{min}^2\right)}\\ &+ \sqrt{\left(\frac{r_{min}^2}{2\left(r_{max}^2-r_{min}^2\right)}\right)^2 + \frac{r_{max}^2}{r_{max}^2-r_{min}^2}u}
\end{split}
\end{equation}
Using the inverse CDF for sampling the radius, then a random point in the North-East plane within the annulus is given by 
\begin{equation}
        N = r\cos\theta, \quad E = r\sin\theta
\end{equation}
where $\theta \sim \mathcal{U}(0,2\pi)$ is uniformly distributed, and 
\begin{equation}
r = r_{\text{min}} + F^{-1}(u)(r_{\text{max}} - r_{\text{min}}).
\end{equation}
The sampling is restricted to the half-annulus by limiting the range of $\theta$.

\subsection{Uniform sampling of a concentric elliptical annulus}\label{samplinghalfellipse}
Similar to the annulus, it is possible to uniformly sample a concentric elliptical annulus avoiding rejection sampling. 

Let $r_{min}$ be the radius of the inner circle, $\theta$ the polar angle measured from the major axis of the ellipse, and $r_{max}(\theta)$ the radius of the ellipse 
\begin{equation}
r_{max}\left(\theta\right) = \frac{ab}{\sqrt{\left(b\cos\theta\right)^2 + \left(a\sin\theta\right)^2}}\label{polarellipse}
\end{equation}
with $a$ and $b$ the major and minor axes. To achieve uniform sampling within the concentric elliptical annulus, the likelihood of sampling $\theta$ should be higher where $r_{max}(\theta) - r_{min}$ is large. This is equivalent to say that the PDF of $\theta$ should~be
\begin{equation}
	f(\theta) d\theta = \frac{dA}{A}
\end{equation}
where $dA$ is the differential area corresponding to the differential angle $d\theta$, and $A$ is the total area of the concentric elliptical annulus.

Given an angle $\theta$, the differential area $dA$ is given by
\begin{align}
	dA &= \frac{1}{2}\left(r_{max}^2(\theta) - r_{min}^2\right)d\theta\label{ellipse_da}
\end{align}
and the total area is computed as
\begin{equation}\label{eq:ellipse_annulus_definition}
	A = \pi (ab - r_{min}^2)
\end{equation}
i.e. the area of the ellipse subtracted the area of the inner circle. Substituting \eqref{polarellipse} into \eqref{ellipse_da} the PDF of the random variable $\theta$ reads
\begin{align}
f(\theta) d\theta &=\frac{1}{2A}\left(
\frac{(ab)^2}{(b\cos\theta)^2 + (a\sin\theta)^2}-r_{min}^2
\right)d\theta
\end{align}
and the resulting CDF is
\begin{equation}
	F(\theta) =  \frac{a b \tan^{-1}\left(\frac{a\tan\theta}{b}\right)-\theta r_{min}^2}{2A}.
\end{equation}

Obtaining a closed form solution of the inverse CDF for the concentric elliptical annulus is infeasible since 
\begin{equation}
	 \frac{a b\tan^{-1}\left(\frac{a\tan\theta}{b}\right)-\theta r_{min}^2}{2A} = u
\end{equation}
has no exact solution for $\theta$. Using a numerical approximation, such as Newton-Raphson, the value of $\theta$ can be approximated. Given a realization $u$ of the uniform distribution $\mathcal{U}$, the approximation is computed as
\begin{equation}
\theta_{k+1} = \theta_k + \frac{f(\theta_k)}{f'(\theta_k)}
\end{equation}
where
\begin{align}
f(\theta_k) &= 	 \frac{a b\tan^{-1}\left(\frac{a\tan\theta_k}{b}\right)-\theta_kr_{min}^2}{2A} -u\\
f'(\theta_k) &= \frac{1}{2A}\left(
\frac{(ab)^2}{(b\cos\theta_k)^2 + (a\sin\theta_k)^2}-r_{min}^2
\right).
\end{align}
Due to the singularity of $\tan(\cdot)$ at $\pm \pi/2$, the uniform distribution is defined in the closed interval $[0,0.24\bar{9}]$. The radius length is sampled in similar fashion as for the annulus.

\subsection{Theoretical sampling performance improvements}\label{sec:sampling_analysis}
Directly sampling a reduced valid subspace leads to a performance improvement proportional to the the ratio of the areas of the original and the reduced subspaces, assuming equal sampling speed. This can be expressed as 
\begin{equation}
    \alpha = \frac{A_{orig}}{A_{valid}}
\end{equation}
where $\alpha$ is the relative performance improvement. Based on this, directly sampling an annulus with inner radius $r_{min}$ and outer radius $r_{max}$ instead of a square with side-length $2r_{max}$ leads to a performance improvement of 
\begin{equation}
    \alpha_{\text{annulus}} = \frac{A_{\text{square}}}{A_{\text{annulus}}} = \frac{4}{\pi \left(1-\left(r_{min}/r_{max}\right)^2\right)}
\end{equation}
considering that $A_{\text{square}} = 4r_{max}^2$ and $A_{\text{annulus}} = \pi (r_{max}^2-r_{min}^2)$. The lower bound is obtained for $r_{min}=0$, leading to a circle within a square, such that $\alpha_{min} = 4/\pi$.
It follows that for a half-annulus, this lower bound doubles.  
For the Informed-RRT*, the performance gain obtained by directly sampling the elliptical half-annulus, instead of the regular informed ellipse, becomes
\begin{equation}
    \alpha_{\text{ell. ann.}} = \frac{A_{\text{ell.}}}{A_{\text{ell. ann.}}} = \frac{c_{best} \sqrt{c_{best}^2 - 4r_{max}^2}}{c_{best} \sqrt{c_{best}^2 - 4r_{max}^2} - 4r_{min}^2}
\end{equation}
given the area of the elliptical annulus in (\ref{eq:ellipse_annulus_definition}).

\begin{figure}[tb]
\centering
    \includegraphics[width=0.45\textwidth]{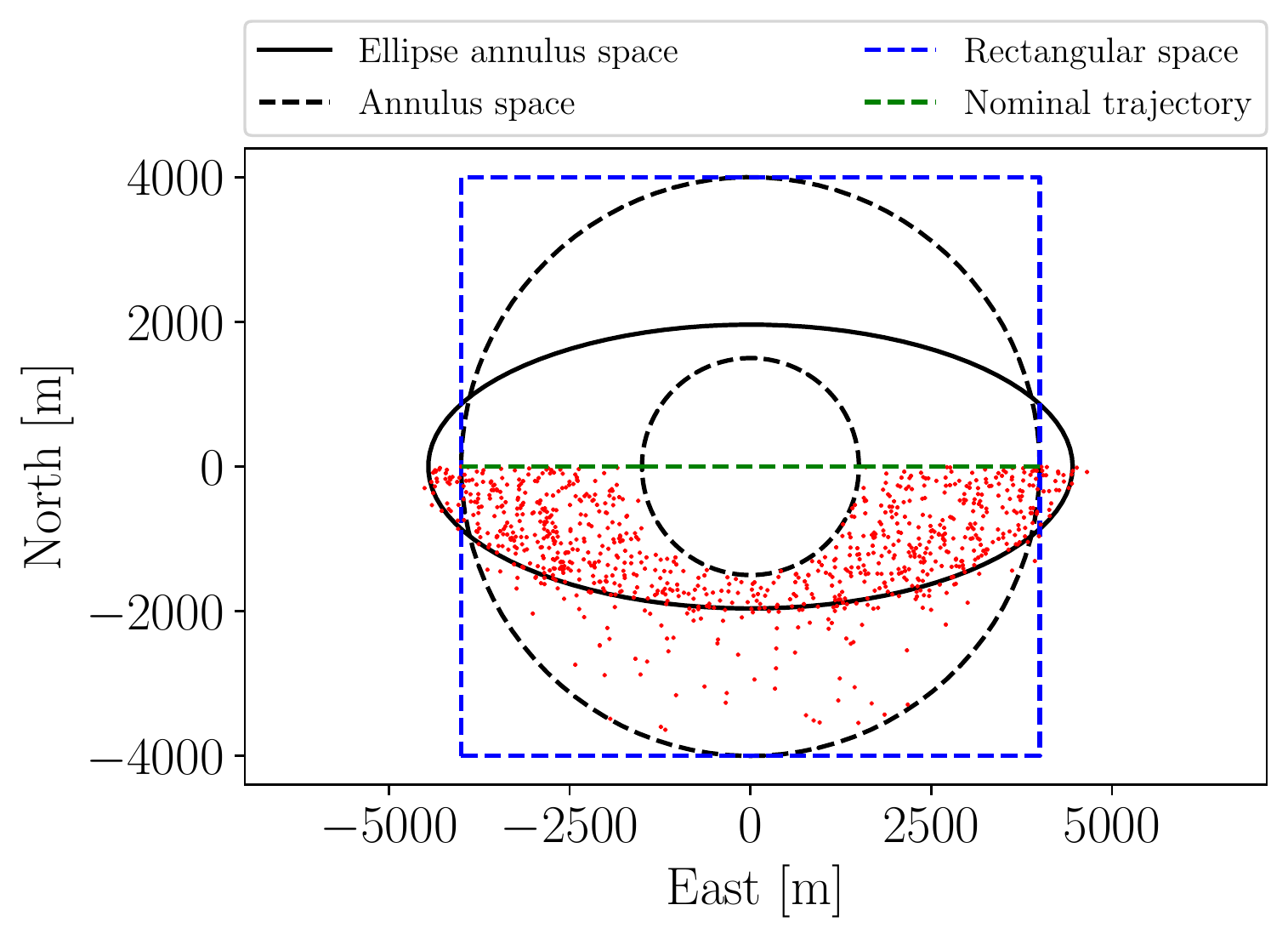}
	\caption{The distribution $n=1000$ samples, where the half-annulus sampling space is decreased into the elliptical half-annulus once an initial feasible path is obtained.}
	\label{fig:sampling_annulus_ellipse}
\end{figure}

The relative performance gain obtained by sampling an elliptical annulus instead of an ellipse thus depends on the inner radius of the annulus and current best cost $c_{best}$. Fig.~\ref{fig:sampling_annulus_ellipse} demonstrates the concentration of samples in the subspace of interest for the collision avoidance case.

\setlength{\textfloatsep}{1pt}
\begin{algorithm}[tb]
\footnotesize
    \LinesNumbered
	\SetKwInOut{Output}{Given}
	\Output{$z_{start}$, $z_{goal}$}
	\SetKw{Given}{$x_{\text{start}}$, $x_{\text{goal}}$}
	$V \leftarrow \{z_{start}\}$,$\quad E \leftarrow \emptyset$,$\quad \mathcal{T} = (V,E)$\;
	\For{\textup{i = $1\dots N$}}{
		\textcolor{red}{$c_{best} \leftarrow \min_{z_{soln}\in Z_{soln}}\{$\texttt{Cost($z_{soln}$)}$\}$\;}\label{infRRTc}
		\textcolor{red}{$z_{rand} \leftarrow$\texttt{Sample($z_{start},z_{goal},c_{best}$)}\;}\label{infRRTs}
		
		$z_{nearest} \leftarrow$ \texttt{NearestNode($\mathcal{T}$,$z_{rand}$)}\;
		$(z_{new},x_{new}) \leftarrow$ \texttt{ExtendTowards($z_{nearest}$,$z_{rand}$)}\;
		\If{\texttt{\upshape Feasible($z_{nearest},z_{new},x_{new}$)}}{
		    $\mathcal{T}\leftarrow$\texttt{InsertNode($\mathcal{T},z_{new}$)}\;
			$Z_{near} \leftarrow$ \texttt{Near($\mathcal{T}$,$z_{new}$,$r$)}\;
			$z_{min} \leftarrow z_{nearest}$\;
			$c_{min} \leftarrow$ \texttt{Cost}($z_{min}$) + \texttt{Line}($z_{min},z_{new}$)\;
			\For{$\forall z_{near} \in Z_{near}$}{
			$c_{new} \leftarrow$ \texttt{Cost}($z_{near}$) + \texttt{Line}($z_{near},z_{new}$)\;
			\If{$c_{new} < c_{min}$}{
			\If{\texttt{\upshape Feasible($z_{near},z_{new},x_{new}$)}}
			{
			    $z_{min}\leftarrow z_{near}$\;
			    $c_{min}\leftarrow c_{new}$\;
			}
			}
			}
			$E\leftarrow E \cup \{z_{min},z_{new}\}$\;
			\For{$\forall z_{near} \in Z_{near}$}{
			$c_{near} \leftarrow$ \texttt{Cost}($z_{near}$)\;
			$c_{new} \leftarrow$ \texttt{Cost}($z_{new}$) + \texttt{Line}($z_{new},z_{near}$)\;
			\If{$c_{new} < c_{near}$}{
			\If{\texttt{\upshape Feasible($z_{new},z_{near},x_{near}$)}}
			{
			    $z_{parent}\leftarrow$ \texttt{Parent($z_{near}$)}\;
			    $E\leftarrow E \backslash \{z_{parent},z_{near}\}$\;
			    $E\leftarrow E \cup \{z_{new},z_{near}\}$\;
			}
			}
			}
		\textcolor{red}{\If{\texttt{\upshape InGoalRegion($z_{new}$)}}
		{
		      $Z_{soln} \leftarrow Z_{soln} \cup \{z_{new}\}$
		    }}\label{infRRTg}
		    
		}
		
		}
		\Return $\mathcal{T}$\;
	\caption{RRT* \textcolor{red}{(Modified sampling)}}\label{mainRRT}
\end{algorithm}
\subsection{Selecting the half-annulus or elliptical half-annulus}\label{samplingchoice}
As described in Section \ref{section_creg_samp_def}, once an initial solution is obtained, the algorithm switches to sampling the elliptical half-annulus, if it promises performance improvement.
For some solution costs $c_{best}$, the area of the elliptical half-annulus may be larger than the half-annulus. A switching condition can be computed based on the area ratio.
Let $A_{he}$ and $A_{sc}$ be the areas of the half ellipse and outer semi-circle, respectively. The elliptical half-annulus will be sampled, rather than the half-annulus, if 
\begin{equation}\label{eq:sampling_test}
\frac{A_{he}}{A_{sc}} < 1 \Leftrightarrow  c_{best} \sqrt{c_{best}^2 - c_{min}^2} < r_{max}^2.
\end{equation}
By squaring \eqref{eq:sampling_test} and solving the associated 4th order polynomial  for $c_{best}$, the following condition $\gamma$ is obtained,
\begin{equation}
\gamma = \sqrt{\frac{c_{min}^2}{2} + \sqrt{\left(\frac{c_{min}^2}{2}\right)^2 + r_{max}^4}}\label{informed_th}
\end{equation}
which is the resulting switching condition for sampling the elliptical half-annulus when $c_{best} < \gamma$.

\section{COLREGs-Informed RRT*}
The underlying planning algorithm consists of a marine-oriented RRT*, which is configured to sample in the COLREGs compliant subset, thus increasing sampling quality and avoiding rejection sampling due to breaching the COLREGs.

\begin{figure*}[!ht]
    \centering
    \begin{subfigure}[b]{0.48\textwidth}
    \centering
    \includegraphics[width=1\linewidth]{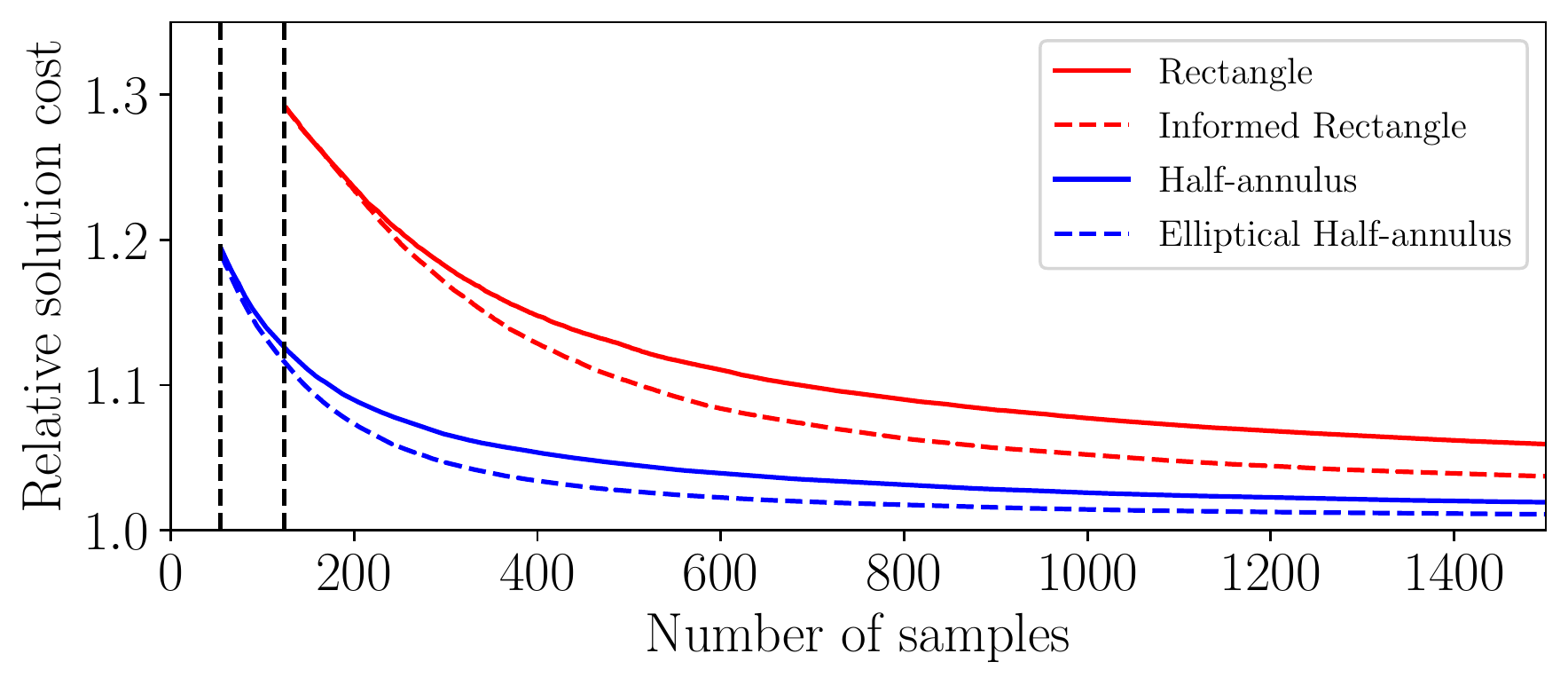}
\caption{Relative solution cost ($c_{best}/c_{min}$). The mean number of samples required to find an initial solution, using the half-annulus and rectangular spaces, is $54$ and $124$ respectively.}\label{perf_cost}
    \end{subfigure}
    \hfill
    \begin{subfigure}[b]{0.48\textwidth}
    \centering
    \includegraphics[width=1\linewidth]{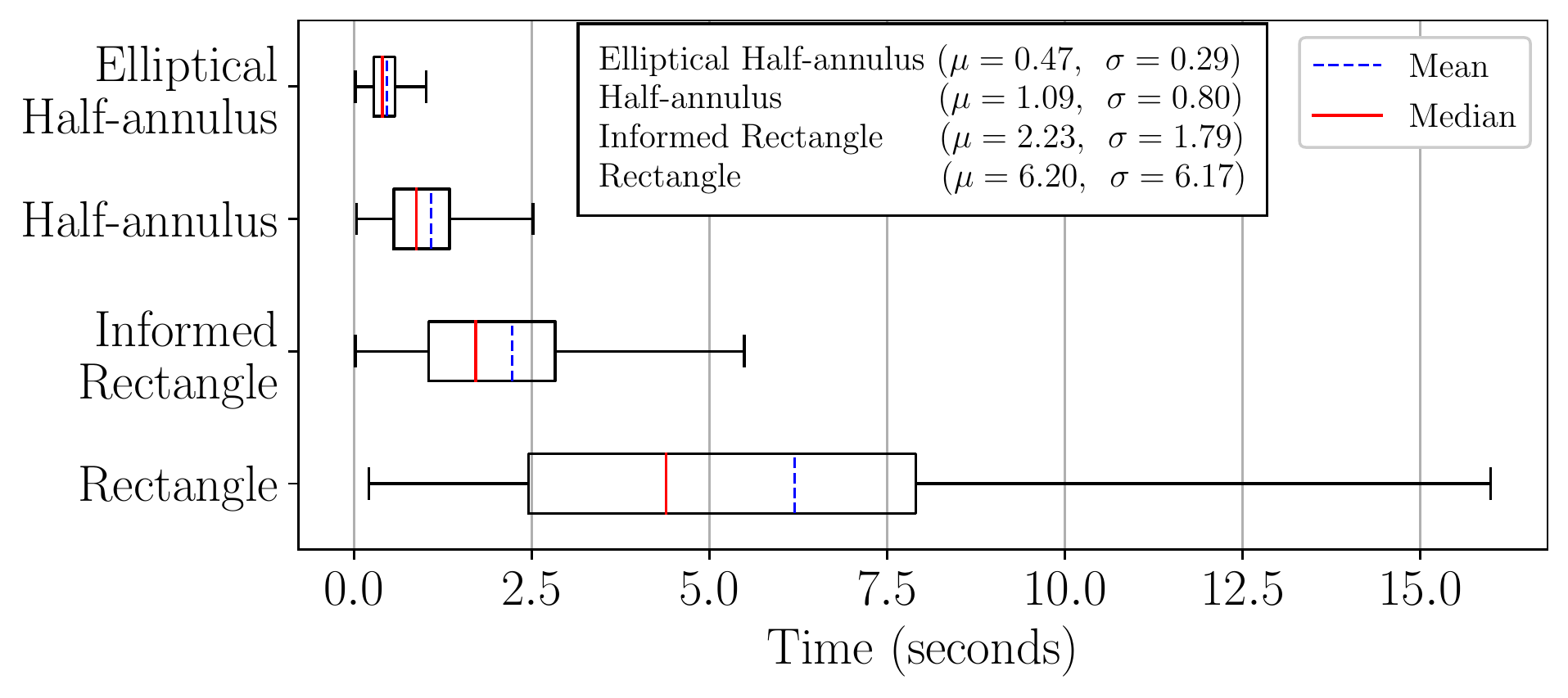}
\caption{Time required to be within 5\% of the optimal solution. The $\mu$ and $\sigma$ values indicate the  mean and standard deviation for the computational time in seconds.}\label{perf_time}
    \end{subfigure}
    \caption{Comparison between the Rectangular space (\cite{karaman2011sampling, Zaccone2020}), Informed Rectangular space (\cite{Gammell2014}) and the two proposed sampling spaces. Each comparison is done over 2500 trials, with the only algorithmic modifications occurring within Line \ref{infRRTs} in Algorithm \ref{mainRRT}, and the utilization of rejection sampling for COLREGs compliance in the two former sampling methods.}\label{perf_comparison}
    \vspace{-0.5cm}
\end{figure*}
\begin{figure}[!ht]
\centering
\includegraphics[width=0.85\linewidth]{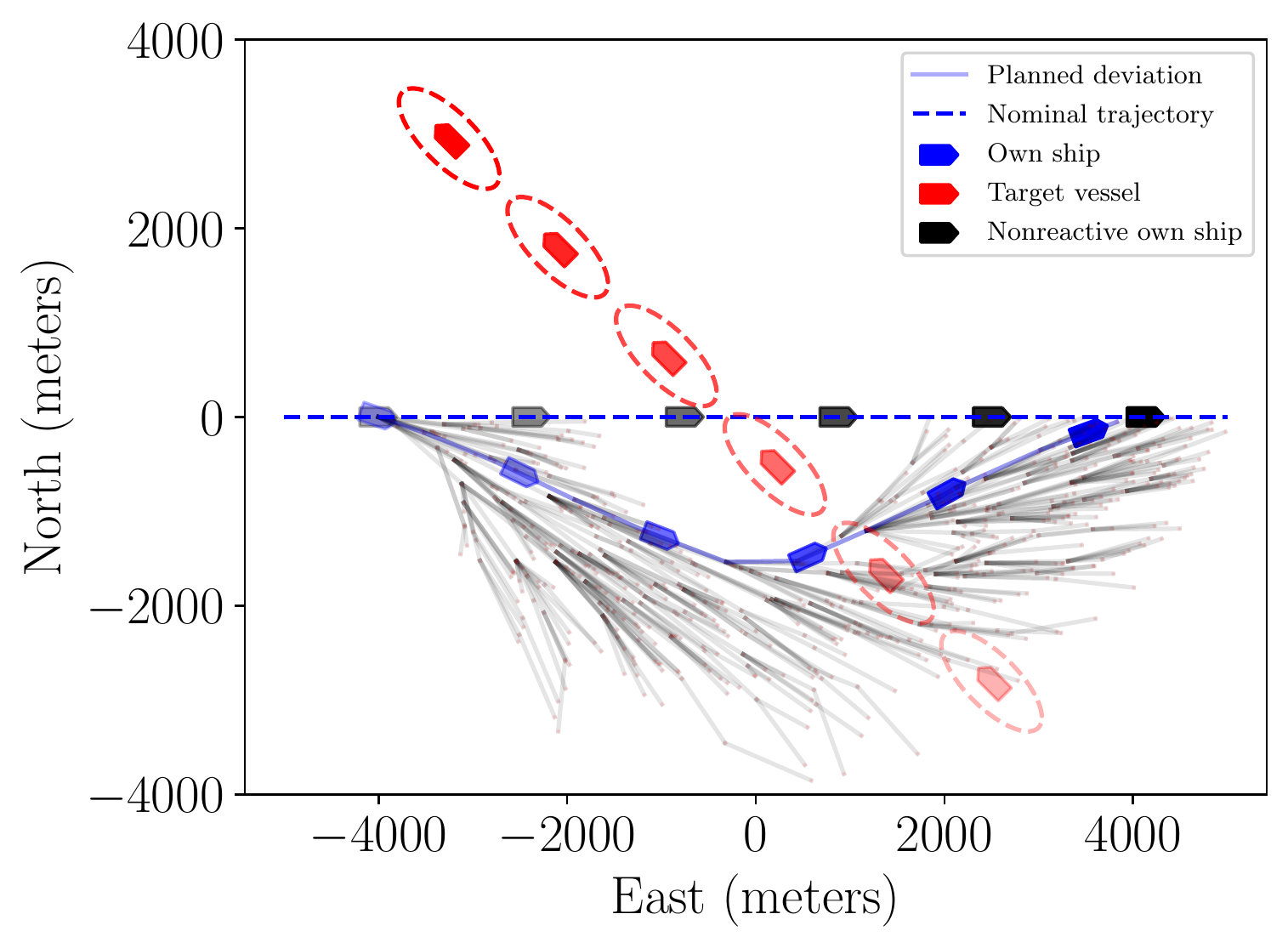}
\caption{Proposed path deviation to a starboard crossing computed by the COLREGs-Informed RRT*}\label{crossing_sim}
\end{figure}

Algorithm~\ref{mainRRT} (adopted from \cite{Gammell2014}) describes the main structure of the marine-oriented RRT* procedure, where the core RRT* components are based on the work presented by \cite{karaman2011sampling}. The method grows a tree $\mathcal{T} = (V,E)$, where $V$ and $E$ represent the sets of nodes and edges, respectively. A given node $z$ represents a North-East position of the vessel at time $t$, with the edges containing relevant cost information. The underlying RRT* method is modified such that nodes resemble waypoints which are compliant with the underlying track-control scheme and the maneuvering capabilities of the vessel. The method \texttt{Feasible($z_{nearest},z_{new},x_{new}$)} is responsible for ensuring validity of the sampled points and the trajectory between two nodes with respect to vessel capabilities (e.g. the turning radius \eqref{turningrad_req}), static obstacles and dynamic obstacles. The red lines (Lines \ref{infRRTc}, \ref{infRRTs} and \ref{infRRTg}) indicate the modifications required for sampling the proposed COLREGs compliant subsets. 
Until an initial solution is obtained, the sampling (\texttt{Sample($z_{start},z_{goal},c_{best}$)}) is solely performed using the proposed method in Section~\ref{samplinghalfannulus}, i.e the half-annulus. Once an initial solution is obtained, the cost is compared to the threshold \eqref{informed_th}, which then initiates the sampling routine described in Section~\ref{samplinghalfellipse}, if  sampling the elliptical half-annulus provides more valuable samples.

\section{Results}
The proposed COLREGs compliant sampling stategies presented in Section \ref{section_creg_sampling} are implemented to yield a marine-oriented RRT* algorithm. Fig.~\ref{crossing_sim} shows a single vessel encounter, where own ship is the give-way vessel in a crossing scenario. The computed path deviation ensures a reaction to the approaching target vessel in ample time, and therefore adheres to Rule 16. Furthermore, the deviation from the nominal trajectory is both COLREGs compliant (w.r.t. Rule 15) and collision free. This is guaranteed by design for both crossing and head-on scenarios for any initial configuration of own ship and target vessel for which $\mathrm{TCPA} > t_{act}$. In an overtaking scenario the computed path is still guaranteed to be COLREGs compliant; however the ability to find a collision free path depends on the difference in forward speed between own ship and target vessel. This is due the chosen definition of the goal region for the RRT* algorithm.

The performance of the two proposed sampling schemes is compared to the baseline RRT* \cite{karaman2011sampling,Zaccone2020} and Informed-RRT* \cite{Gammell2014} algorithms, which use rectangular sampling spaces and rejection sampling to adhere to the COLREGs. The methods are evaluated in terms of the obtained solution cost after $n$ samples, number of iterations required to obtain an initial feasible solution, and overall execution speed. 

As pointed out by the theoretical analysis in Section~\ref{sec:sampling_analysis}, an increased performance is obtained by sampling directly in the COLREGs compliant subspace. Sampling in the half-annulus, compared to the rectangular space, obtains an initial path in a shorter amount of time (proportional to the area difference). Fig. \ref{perf_cost} shows the relative solution cost, as well as the average amount of samples required to find an initial solution. The proposed COLREGs compliant subspaces achieve a higher sampling density in the relevant regions and thus find an initial solution with fewer iterations. Fig.~\ref{perf_time} compares the computational speed of RRT* for different sampling strategies, and it is evident the clear computational advantage offered by the proposed sampling methods. 

\section{Conclusions}
The paper presented the concept of COLREGs compliant sampling spaces for use within the RRT* algorithm to generate collision free path alterations for vessels in close quarters. The proposed sampling spaces, i.e. the half-annulus and elliptical half-annulus, directly encode the COLREGs rules 13-15; hence the RRT* algorithm searches for path deviations that are guaranteed to adhere the rules of safe navigation. This results in a significant enhancement of the path planner performance when compared to reference sampling strategies with respect to the metrics of mean time to discover a feasible path, computational speed to find the optimal solution and convergence to the optimal cost. This improvement is due to the specialization of the sampling space, which both increases the probability of gaining value for each sample and removes the need of rejection sampling. These high-gain sampling strategies could prove valuable when extending the implementation to include additional complexities, such as system dynamics, since it will result in fewer solutions to boundary-value problems.

\addtolength{\textheight}{-2.5cm}   


\bibliographystyle{IEEEtran}
\bibliography{mainbib.bib}
\end{document}